%% file: neurips_2023.tex
\documentclass{article}

     \PassOptionsToPackage{numbers, compress}{natbib}

\usepackage[preprint]{neurips_2023}
\usepackage{graphicx}

\usepackage[utf8]{inputenc} 
\usepackage[T1]{fontenc}    
\usepackage{hyperref}       
\usepackage{url}            
\usepackage{booktabs}       
\usepackage{amsfonts}       
\usepackage{nicefrac}       
\usepackage{microtype}      
\usepackage{xcolor}         
\usepackage{xspace}
\usepackage{multirow}
\newcommand{\baby}{\textsc{RecurrentGPT}\xspace}

\usepackage{todonotes}

\usepackage{appendix}

\title{\baby: \\
Interactive Generation of (Arbitrarily) Long Text}

\newcommand{\ethz}{\includegraphics[width=1em,height=1em]{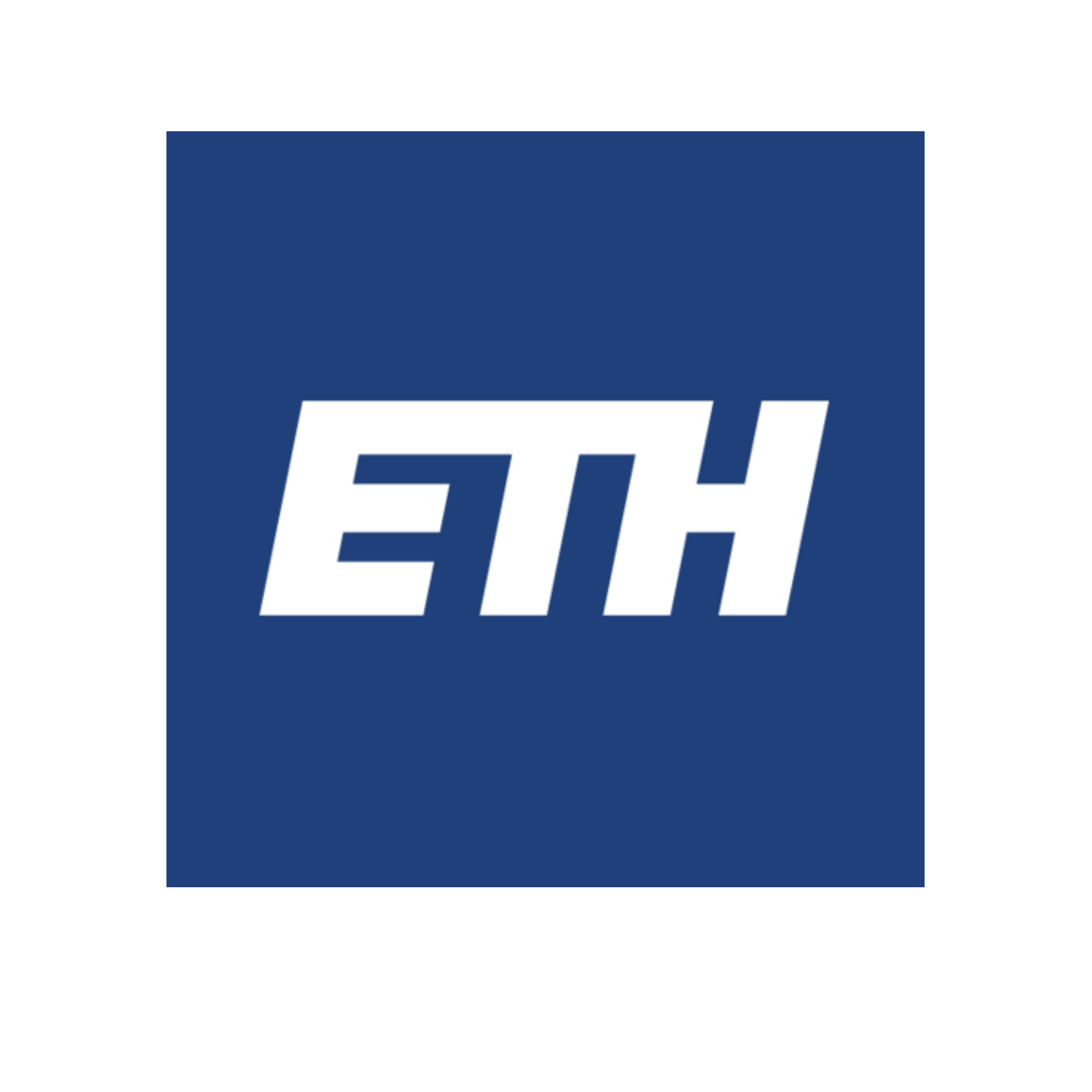}}

\author{
  Wangchunshu Zhou{\thanks{Equal Contribution}*\ethz}~\ 
  Yuchen Eleanor Jiang{*\ethz}~\ Peng Cui{\ethz}~\ Tiannan Wang\\
  \textbf{Zhenxin Xiao~\ 
  Yifan Hou{\ethz}~\ 
  Ryan Cotterell{\ethz}~\ 
  Mrinmaya Sachan{\ethz}}\\
  {\ethz}ETH Z\"{u}rich \\
\texttt{\{\href{mailto:wangchunshu.zhou@inf.ethz.ch}{wangchunshu.zhou},
\href{mailto:yuchen.jiang@inf.ethz.ch}{yuchen.jiang},
\href{mailto:peng.cui@inf.ethz.ch}{peng.cui}\}@inf.ethz.ch} \\
\texttt{\href{mailto:hugothebestwang@gmail.com}{hugothebestwang@gmail.com}, \href{mailto:alanshawzju@gmail.com}{alanshawzju@gmail.com}}
\\
\texttt{\{\href{mailto:yifan.hou@inf.ethz.ch}{yifan.hou}, \href{mailto:ryan.cotterell@inf.ethz.ch}{ryan.cotterell}, \href{mailto:mrinmaya.sachan@inf.ethz.ch}{mrinmaya.sachan}\}@inf.ethz.ch}
}

\begin{document}

\maketitle

\input{sections/0_abstract}
\input{sections/1_introduction}

\input{sections/2_RecurrentGPT}
\input{sections/3_Experiments}

\input{sections/5_RelatedWorks}
\input{sections/6_Limitations}
\input{sections/7_Conclusion}

\bibliography{neurips_2023}
\bibliographystyle{unsrtnat}
\input{sections/appendix}

\end{document}

%% file: sections/0_abstract.tex
\begin{abstract}
The fixed-size context of Transformer makes GPT models incapable of generating arbitrarily long text. In this paper, we introduce \baby, a language-based simulacrum of the recurrence mechanism in RNNs.
\baby is built upon a large language model (LLM) such as ChatGPT and uses natural language to simulate the Long Short-Term Memory mechanism in an LSTM. At each timestep, \baby generates a paragraph of text and updates its language-based long-short term memory stored on the hard drive and the prompt, respectively. This recurrence mechanism enables \baby to generate texts of arbitrary length without forgetting. Since human users can easily observe and edit the natural language memories, \baby is interpretable and enables interactive generation of long text. \baby is an initial step towards next-generation computer-assisted writing systems beyond local editing suggestions. 
In addition to producing AI-generated content (AIGC), we also demonstrate the possibility of using \baby as an interactive fiction that directly interacts with consumers. We call this usage of generative models by ``\textbf{AI} \textbf{a}s \textbf{C}ontents'' (\textbf{AIAC}), which we believe is the next form of conventional AIGC.
We further demonstrate the possibility of using \baby to create personalized interactive fiction that directly interacts with readers instead of interacting with writers.
More broadly, \baby demonstrates the utility of borrowing ideas from
popular model designs in cognitive science and deep learning for prompting LLMs.
Our code is available at \url{https://github.com/aiwaves-cn/RecurrentGPT} and an online demo is available at \url{https://www.aiwaves.org/recurrentgpt}.
\end{abstract}

%% file: sections/1_introduction.tex
\section{Introduction}

Large Language Models (LLMs)~\citep{gpt,gpt2,gpt3,gpt3.5,gpt4} such as ChatGPT have proven to be highly effective tools for assisting with various routine writing tasks, including emails and blog posts. Nevertheless, due to the fixed-size context design inherent in the Transformer~\citep{transformer} architecture, it is unfeasible to generate long texts (e.g., novels) solely by prompting LLMs. In contrast, recurrent neural networks (RNNs)~\citep{RNN, hochreiter1997long}, in theory, possess the capacity to generate sequences of arbitrary length, thanks to their recurrence mechanism: RNNs maintain a hidden state that undergoes updates at each time step, employing the current time step's output as the input for the subsequent time step. In practice, however, RNNs suffer from the problem of vanishing and exploding gradients and are hard to scale up.

To this end, a number of works~\citep{dai*2019transformerxl,Rae2020Compressive,bulatov2022recurrent} attempt to equip Transformers with an RNN-like recurrence mechanism. While achieving promising results on long text modeling and generation, these recurrence-augmented Transformers require substantial architectural modifications that have not been proven to scale well.
The majority of current LLMs continue to employ the original Transformer architecture with minimal alterations.

\begin{figure}[t]
    \centering
   \includegraphics[width=\textwidth,trim={0cm 0cm 0cm 0cm} ,clip]{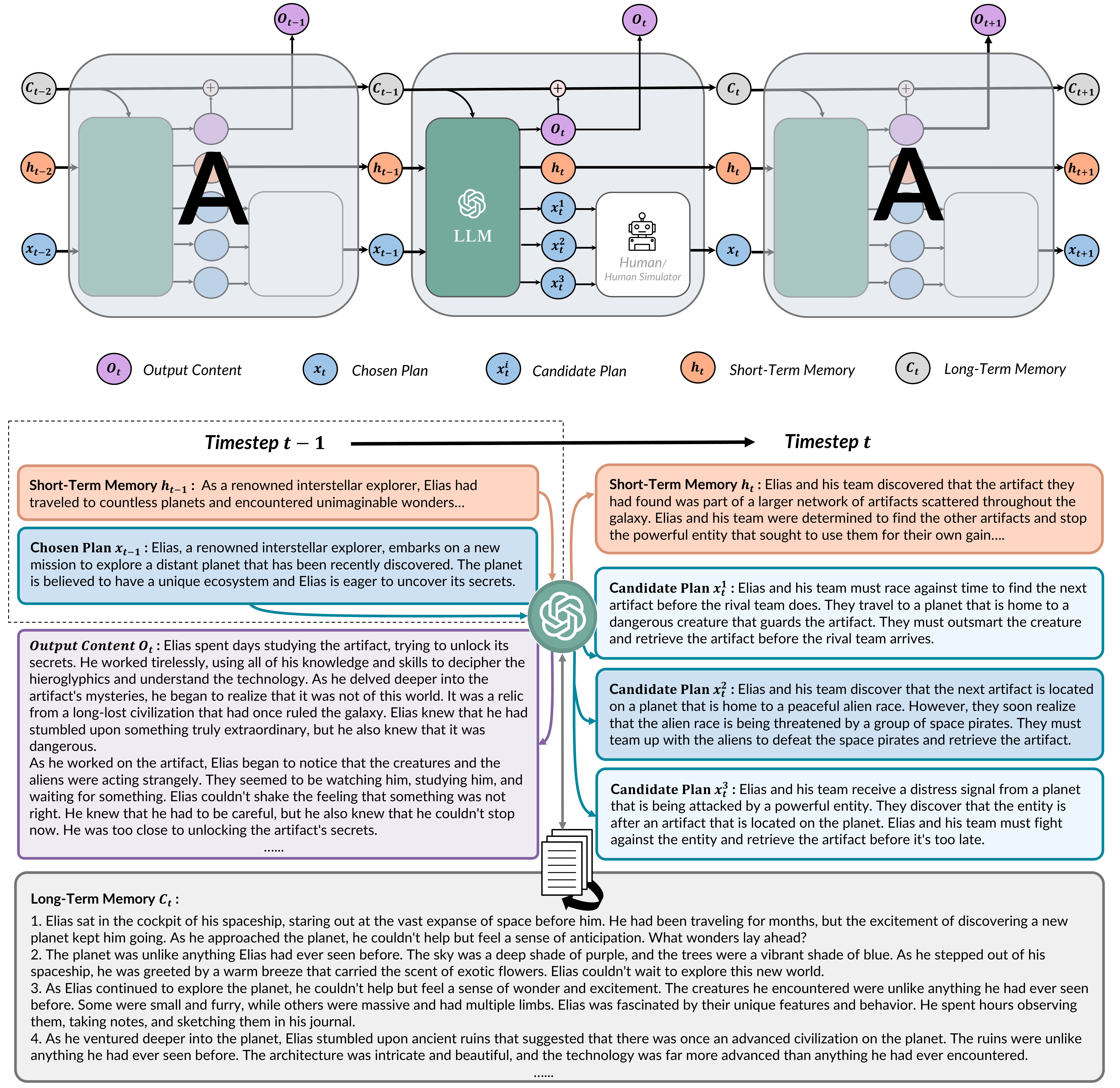}
    \caption{Illustration of the \baby framework. \baby enables recurrent prompting with LLMs by simulating an RNN using natural language building blocks and defines the recurrent computation graph with prompts. \looseness=-1}
    \label{fig:main}
\end{figure}

In this paper, we introduce \baby, a language-based simulacrum of the recurrence mechanism in RNNs. 
As illustrated in Figure \ref{fig:main}, \baby replaces the vectorized elements (i.e., cell state, hidden state, input, and output) in a Long-short Term Memory RNN (LSTM)~\citep{hochreiter1997long} with natural language (i.e., paragraphs of texts), and simulates the recurrence mechanism with prompt engineering. At each timestep $t$, \baby receives a paragraph of text and a brief plan of the next paragraph, which are both generated in step $t-1$. It then attends to the long-term memory, which contains the summaries of all previously generated paragraphs and can be stored on hard drives, and relevant paragraphs can be retrieved with semantic search. \baby also maintains a short-term memory that summarizes key information within recent timesteps in natural language and is updated at each time step. \baby combines all aforementioned inputs in a prompt and asks the backbone LLM to generate a new paragraph, a short plan for the next paragraph, and updates the long-short term memory by rewriting the short-term memory and appending the summary of the output paragraph to the long-term memory. These components are then re-used in the next time step, resulting in a recurrence mechanism for the generation process.
With the language-based recurrence mechanism, \baby alleviates the need for any architectural modification and can be integrated into any powerful LLM, making it capable of generating arbitrarily long text beyond the fixed-size context window.

In addition to surpassing the fixed-size context limitation, \baby enhances the interpretability of the recurrence mechanism in comparison to the vector-based recurrence mechanism employed in RNNs. This improvement stems from the ability to observe the specific segments of long-term memory that are attended to, as well as the manner in which short-term memory is updated, through a simple examination.
More importantly, employing natural language as building blocks enables human engagement with \baby, allowing for the human manipulation of its memories and plans for future generations. 
Human interaction also prevents \baby from deviating from desired behavior, a challenge commonly encountered with recent autonomous GPT-based agents such as AutoGPT\footnote{\url{https://github.com/Significant-Gravitas/Auto-GPT}}. Given that current state-of-the-art computer-assisted writing systems~\citep{coauthor,10.1145/3544548.3580969} primarily focus on localized editing suggestions and treat LLMs as black-boxes, we believe \baby represents a step towards next-generation computer-assisted writing systems for interactive long text generation that also offer interpretability.

We then extend the utilization of \baby beyond its role as a tool for producing AI-generated content (AIGC) by exploring its potential for direct interaction with consumers, rather than solely with content creators. Specifically, we convert \baby to a personalized interactive fiction wherein it generates multiple prospective plans for the subsequent actions, allowing players to choose and explore the one that captures their interest. Moreover, in addition to selecting from model-generated plans, players possess the capability to devise their own plans. Such a capacity is unattainable within conventional interactive fictions, as the narratives and options are conventionally predetermined. We denote this new paradigm as ``AI As Content'', signifying the utilization of generative AI as a medium that actively interacts with consumers, instead of being confined to the role of a mere tool for content creators. 
Through \baby, we perceive a preliminary stride towards a future where AI models will eventually become collaborative partners in our creative endeavors.

In our experiments, we build \baby upon ChatGPT and find that exhibits the capability to autonomously generate remarkably extensive texts, spanning thousands of tokens, while maintaining both coherency and engagement. In stark contrast, vanilla ChatGPT is constrained to generating a few hundred of tokens before encountering issues such as repetitive content or a decline in coherence.%
Moreover, \baby can help human writers produce arbitrarily long text with ease, reducing much of the human efforts required for writing long creative texts such as novels. The contributions of this paper can be summarized as follows:
\begin{itemize}
    \item We propose \baby, a language-based simulacrum of the recurrence mechanism in RNNs that mitigates the fixed-size context limitation of LLMs such as ChatGPT.
    \item We show that \baby can generate very long texts either on its own or serve as an interactive writing assistant, helping human writers write arbitrarily long texts.
    \item We introduce a new use case of generative AI that uses generative models to directly interact with consumers of text, as opposed to the conventional practice that uses them as tools for content creation, by using \baby as a personalized interactive fiction for content curation.
\end{itemize}
Furthermore, it is important to underscore that \baby illustrates the possibility of drawing inspiration from well-established model designs in the fields of cognitive science and deep learning, with the aim of generating long form text via prompting of LLMs.


%% file: sections/2_RecurrentGPT.tex
\section{\baby}

We describe \baby in detail in this section. \baby is a natural language-based counterpart of the recurrence mechanism in RNNs. \baby simulates an LSTM by (1) modeling all vector-based components in an LSTM, including input vectors $x_t$, output vectors $y_t$, hidden states $h_t$, and cell states $c_t$, with natural language; (2) modeling the recurrent computation graph in an LSTM with natural language prompts, and (3) replacing the trainable parameters in RNNs by a frozen LLM. In theory, the backbone of \baby can be any LLM or text-to-text model, we opt for ChatGPT because of its capability and popularity.

Formally, we define \baby as a computational function parametrized by an LLM with parameter $\theta$ and a prompt template $\mathcal{P}$. Recall that the recurrent computation graph of an LSTM can be summarized as:
\begin{equation}
    o_{t+1}, h_{t+1}, c_{t+1} = \textsc{LSTM}(x_{t+1}, h_{t}, c_{t}, \theta)
\end{equation}
where $\theta$ denotes the model parameters, $x_{t+1}$ equals to $o_{t}$, and $h_{t}, c_{t}$ are the long/short-term memories at timestep $t$, respectively. 

By analogy, the recurrence mechanism in our model can be expressed by:
\begin{equation}
    o_{t+1}, x_{t+1}, h_{t+1}, c_{t+1} = \baby(o_{t}, x_{t}, h_{t}, c_{t}, \theta, \mathcal{P})
\end{equation}
where $o_{t}, x_{t}, h_{t},$ and $c_{t}$ denote the natural language-based building blocks including content, plan, short-term memory, and long-term memory, at time step $t$, respectively. Here $x_{t+1}$ does not equal $o_{t}$ and is instead separately generated, which is different from conventional RNNs.
We first describe each building block in \baby and then present how our prompt $\mathcal{P}$ enables \baby to recurrently generate arbitrarily long texts.


\subsection{Language-based Building Blocks}

\paragraph{Input/Output} The input and output of \baby at each timestep include a paragraph of text that gets appended to the final text produced and an outline for the next paragraph to be generated. We refer to these two as the ``content'' and ``plan'', respectively. As illustrated in Figure \ref{fig:main}, contents typically consist of 200-400 words and should be mostly ready for reading. Whereas plans are outlines for the next content and typically consist of 3-5 sentences. At each timestep, the content and plan generated in the previous timestep are used as input to \baby, allowing recurrent computation. \baby is designed to produce plans in addition to contents as allowing users to read and edit plans increases interpretability and facilitates human-computer interaction.

\paragraph{Long-Short Term Memory} Similar to an LSTM, \baby maintains long-short term memory across timesteps. As illustrated in Figure \ref{fig:main}, long-term memory summarizes all previously generated contents to minimize information lost when generating long texts. Since the generated content can be arbitrarily long and cannot fit in the context size of LLMs, we implement the long-term memory in \baby with a VectorDB approach by embedding the content generated in each timestep with sentence-transformers~\citep{reimers-2019-sentence-bert}. This approach enables \baby to store even longer memory compared to previous memory-based Transformers~\citep{dai*2019transformerxl,bulatov2022recurrent} as it can store memory in disk space instead of GPU memory. This can be important in several use cases where the users may not have high-end GPUs in their devices.

Short-term memory, on the other hand, is a short paragraph of texts summarizing key information across recent timesteps. The length of the short-term memory is controlled to 10-20 sentences so that it can fit into the prompt and can be updated by the LLM backbone. By combining long-short term memory, \baby can maintain coherence with recently generated content and also recall key information that was generated long before. This is impossible with vanilla LLMs because they can only take a few previously generated texts in the input.

\baby can be initialized using a simple prompt that instructs the LLM to generate the aforementioned components with texts specifying the topic of the novel and other background information. When using \baby to continue writing a novel, users can write down (or prompt ChatGPT to generate) a short-term memory and an initial plan.

\subsection{Language-based Recurrent Computation}
While RNNs achieve recurrent computation by implementing a feedback loop in the computation graph, \baby relies on prompt engineering to simulate the recurrent computation scheme. As illustrated in Figure \ref{fig:main}, \baby simulates the computation graph in RNNs with a prompt template, which is presented in Figure 1 in the Appendix, and some simple Python code\footnote{We present the prompt in Appendix \ref{app:prompt} due to space constraints.}. 

At each timestep, \baby constructs the input prompts by filling the prompt template with input content/plan and its internal long-short term memory. In particular, since the long-term memory cannot fit into the context size, we use the input plan as the query to perform a semantic search over the VectorDB-based long-term memory and fit a few most relevant contents into the prompt. The prompt then instructs the LLM backbone to generate new contents, plans, and updated short-term memory. As illustrated in Figure 1 in the Appendix, our prompt encourages the LLM to update the short-term memory by discarding information that is no longer relevant and adding useful new information while maintaining its length within a range so that it can always fit in the context size. 
It is noteworthy that we prompt the LLM to generate multiple (e.g., 3 in our experiments) plans. This improves the diversity of outputs and makes human-computer interaction more friendly by allowing human users to select the most suitable plan. We also give users the option to write plans on their own if none of the generated plans is desirable. To make \baby capable of generating long texts autonomously without human intervention, we add a prompt-based human simulator to select a good plan and revise it for the next timestep.

\subsection{Interactive Long Text Generation with \baby}

While \baby can generate long texts on its own with the recurrence mechanism, its language-based computation scheme offers unique interpretability and interactivity. Compared to conventional computer-assisted writing systems that use language models as black boxes and only give next phrase/sentence suggestions, \baby enjoys the following advantages:
\begin{itemize}
    \item It is more efficient at reducing human labor because it makes paragraph/chapter-level progresses instead of local writing suggestions.
    \item It is interpretable because users can directly observe its language-based internal states.
    \item It is interactive because humans can edit their building blocks with natural language.
    \item It is customizable because users can easily modify the prompts to customize the model according to their own interests (e.g., the style of output texts, how much progress to make for each timestep, etc.)
\end{itemize}
In addition, human interaction can also help correct accidental mistakes made by \baby when autonomously generating long texts and prevent error propagation, which is a major bottleneck for long text generation. 

%% file: sections/3_Experiments.tex
\section{Experiments}

\subsection{Experimental Settings}

\paragraph{Tasks} We test the empirical effectiveness of \baby in this section. In particular, we evaluate \baby in three different settings including:
\begin{itemize}
    \item Autonomously generating long texts without human interaction.
    \item Collaboratively generating long texts with a human writer
    \item Directly interacting with text consumers as interactive fictions.
\end{itemize} 

In each of these tasks, we test with a diverse set of genres of novels including science fiction, romance, fantasy, horror, mystery, and thriller novels. To test the effectiveness of \baby for texts of different length, we generate novels of medium length ($\sim$ 3000 words) for horror, mystery, and thriller, and generate longer novels ($\sim$ 6000 words) for sci-fi, romance, and fantasy.

\paragraph{Baselines} Although \baby is the first work on using LLMs to generate arbitrarily long texts, we can still compare it against some reasonable baselines and ablated variants, as listed below:
\begin{itemize}
    \item \textbf{Rolling-ChatGPT}, a simple baseline that prompts ChatGPT to start writing a novel given a genre of literature and some outlines or background settings, and then iteratively prompts ChatGPT to continue writing after reaching the context length limit. This baseline is roughly equivalent to using a sliding context window trick for generating long texts with Transformers.
    \item \textbf{RE$^{3}$}~\citep{yang-etal-2022-re3} is a hierarchical long story generation baseline that first prompts an LLM to generate an outline for the story and then generates the story following the outline with some re-ranking and re-writing pipelines. We re-implement it with ChatGPT to ensure a fair comparison.
    \item \textbf{DOC}~\citep{yang2022doc} is the state-of-the-art long story generation baseline that improves \textbf{RE$^{3}$} with outline control. We re-implement DOC by replacing OPT-175B~\citep{zhang2022opt} with ChatGPT and removing the detailed controller, which is impossible to use because we do not have access to ChatGPT weights. In general, we find that our re-implementation results in slightly better quality because of the improvement on the backbone LLM.
\end{itemize}

It's noteworthy that in principle, both the baselines can not generate arbitrarily long texts while remaining coherent. This is because the \textbf{Rolling-ChatGPT} baseline forgets previously generated contents very quickly. On the other hand, \textbf{RE$^{3}$} and \textbf{DOC} fixes the outline in the first stage, which limits the overall length of the story to be generated.

\input{tabs/table_main}

\paragraph{Evaluation Metrics}

For evaluation, we follow \citet{yang-etal-2022-re3} and conduct a human evaluation by comparing \baby with the baselines according to two dimensions: 
\begin{itemize}
    \item \textbf{Interesting}: How interesting are the generated novels for common readers?
    \item \textbf{Coherent:} How well are the paragraphs organized and connected with each other?
\end{itemize} 

We omit the ``quality'' or ``humanlike'' metrics following~\citet{yang2022doc} since all baselines are based on ChatGPT which can produce high-quality texts most of the time.  We evaluate the compared models by pairwise comparison. Specifically, we give two novels (A and B, with random order) generated by different compared methods to human annotators with good English proficiency and instruct them to label whether novel A or novel B is better, or they are indistinguishable, in terms of interestingness and coherence. Following the human evaluation settings in~\citet{yang2022doc}, we sample 20 generated novels for each genre and assign 3 annotators for each novel. 

\subsection{Results}

As shown in Table \ref{tab:main}, we find that \baby is favored by human readers for both interestingness and coherence with a relatively large margin compared to both the rolling-window baseline and prior state-of-the-arts like RE$^3$ and DOC. This confirms our intuition that recurrent computation is important for long text generation. The gap is larger for longer novels, which confirms the advantage of \baby on generating very long texts. Finally, human annotators prefer \baby in all novel genres. This confirms its robustness on different types of long texts. 

To better understand the effectiveness of \baby, we also conduct an ablation study by comparing \baby with with ablated variants without either short-term or long-term memory, and the variant that uses GPT-4 as the backbone model. The results are shown in Table \ref{tab:ablation}. We can see that long/short-term memory mainly contributes to the coherence of generated texts, which correlates well with our intuition. \baby with GPT-4 as the backbone LLM is drastically favored compared to its counterpart using ChatGPT/GPT-3.5-turbo. This confirms the potential of \baby when equipped with more powerful LLMs. We present a few sample novels generated by \baby in the Appendix for qualitative evaluation.

\begin{figure}[t]
    \centering
   \includegraphics[width=\textwidth,trim={0cm 0cm 0cm 0cm} ,clip]{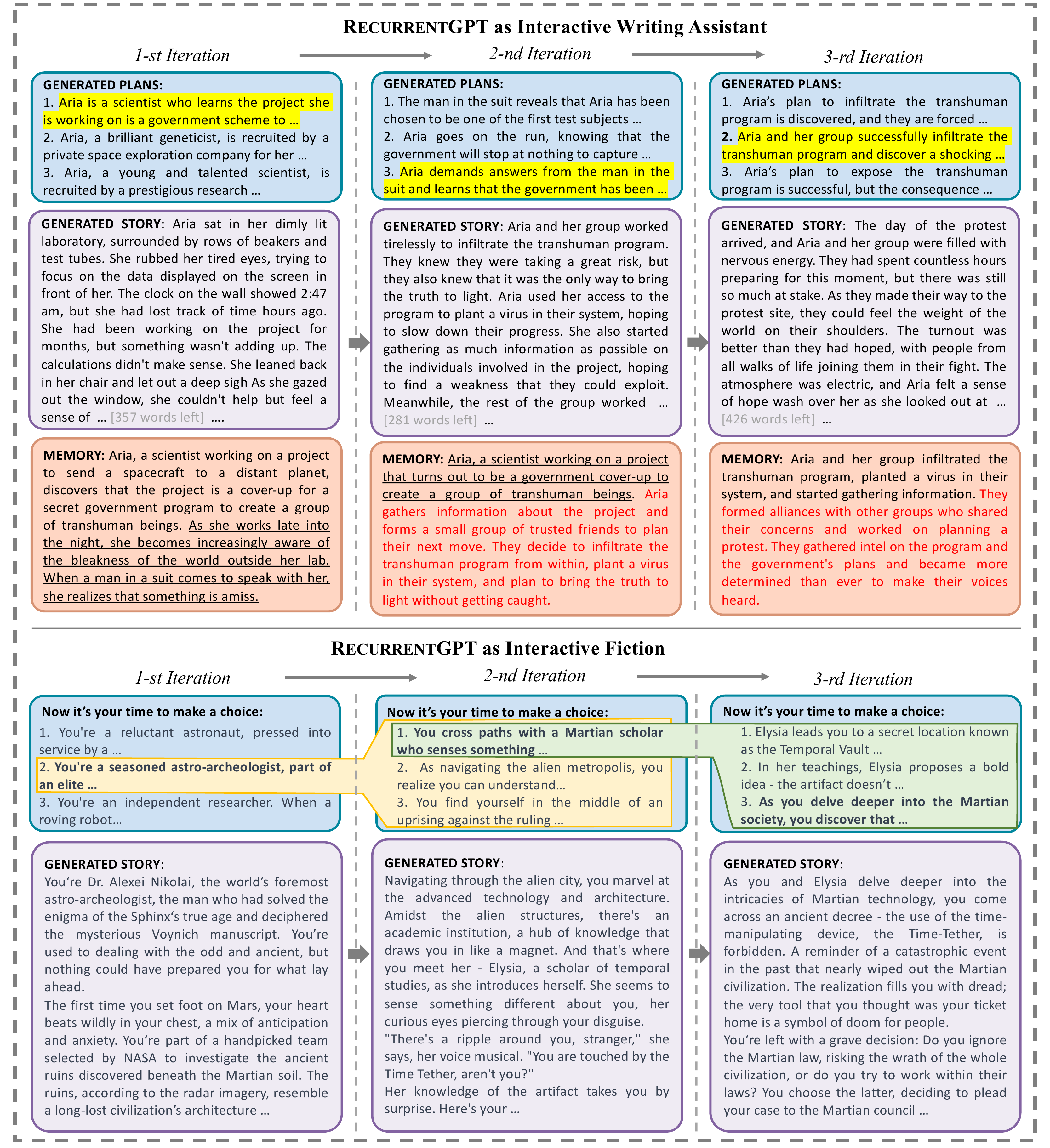}
     \vspace{-6pt}
    \caption{Qualitative analysis of using \baby as an interactive writing assistant and an interactive fiction. Highlighted plans or choices are that selected by human users.}
     \vspace{-15pt}
    \label{fig:case}
\end{figure}

\input{tabs/tab_ablation}

\subsection{\baby as Interactive Writing Assistant}

We then test the usefulness of \baby as an interactive writing assistant from a human-AI interaction perspective.  As illustrated in Figure \ref{fig:case}, a human writer starts by choosing the topic he/she wants to write and writes a short paragraph describing the background and the outline of the book. Then \baby automatically generates the first paragraphs and provides a few possible options for the writer to continue the story. The writer may select one from them and edit it if needed. He or she can also write a short plan for the next few paragraphs by him/herself if generated plans are all inappropriate, which makes human-AI co-writing process more flexible. We show a Gradio\footnote{https://gradio.app/}-based interface that allows human writers to write different genres of novels by interacting with \baby in Appendix \ref{app:demo}.

According to a small-scale human user study, \baby significantly improves the productivity of human writers\footnote{We will conduct a larger-scale user study and present the details and results in the revised version.}, and the improvements mainly come from: (1) reducing the time for typing long texts by writing or choosing short plans and letting \baby generate the actual texts; and (2) reducing the time for designing less important plots by selecting plans from \baby generated ones, according to user feedback. Moreover, users feel that \baby is more interpretable and controllable compared to conventional AI writing assistants that act as black-boxes since the language-based components in \baby are transparent and editable for users. Finally, compared to the previous methods that hierarchically generate long texts such as DOC and RE$^3$, human users prefer our system since iteratively and interactively writing long texts is more flexible and controllable. Finally, our system is very different from most existing AI writing assistants since they focus on providing local writing suggestions within phrases or a few sentences, whereas \baby can generate a few paragraphs at a time.

\subsection{\baby as Interactive Fiction}

We also test the possibility of using \baby as personalized interactive fiction. This use case is very similar to \baby as AI writing assistants. The main differences are two-fold as illustrated in Figure \ref{fig:case}: (1) the shift from the third-person perspective to the first-person perspective, which aims to foster a sense of immersion for human players, and (2) making \baby generate plans that involve important choices for the main character as opposed to general plans for the next paragraphs. The adaptation can be easily implemented by slightly modifying the prompt.

Our user study shows that \baby can interact with human players and directly provide content of good quality for human consumers. Human players also find the possibility of writing free-form texts as their actions in interactive fiction largely improve their interestingness. This confirms the potential of directly using generative AI as content, instead of using them as tools to produce content. However, we also find that \baby sometimes produces less consistent content and low-quality options that are not very relevant or reasonable. We believe this can be improved by using a more powerful LLM backbone, fine-tuning the LLM backbone with supervised fine-tuning or reinforcement learning from human feedback, or designing better prompts. We leave this for future work.

%% file: tabs/table_main.tex
\begin{table}[h]
\centering
\caption{Pair-wise comparison of \baby with baselines for 20 novels of different genres. Results in different comparisons are not comparable with each other. Bold indicates significance with $p < 0.05$.}
\resizebox{1 \textwidth}{!}{
\begin{tabular}{lcccccc}
\toprule

Novel genres & \multicolumn{2}{c}{Sci-fi} & \multicolumn{2}{c}{Romance}  & \multicolumn{2}{c}{Fantasy}  \\
$\sim$ 6000 words & Interesting $\uparrow$ & Coherent $\uparrow$ & Interesting $\uparrow$ & Coherent $\uparrow$ & Interesting $\uparrow$ & Coherent $\uparrow$ \\
\midrule
\baby & \bf 94.7 & \bf 86.5  &  \bf 91.4 & \bf 84.8 & \bf 95.9 & \bf 85.1 \\
Rolling-ChatGPT & 7.8 & 14.3 & 9.0 & 18.2 & 6.5 & 13.7 \\
\midrule
\baby & \bf 68.3 & \bf 65.7 & \bf 71.4 & \bf 69.2 & \bf 63.8 & \bf 62.0 \\
RE$^{3}$ & 31.9 & 28.5 & 28.1 & 25.3 & 35.1 & 33.8 \\
\midrule
\baby & \bf 66.1 & \bf 59.3  &  \bf 77.2 & \bf 63.4 & \bf \bf 61.0  & \bf 56.5 \\
 DOC & 30.7 & 38.1 & 25.3 & 29.8 & 31.2 & 40.3\\
\midrule
\midrule
Novel genres  & \multicolumn{2}{c}{Horror} & \multicolumn{2}{c}{Mystery} &  \multicolumn{2}{c}{Thriller} \\
$\sim$ 3000 words & Interesting $\uparrow$ & Coherent $\uparrow$ & Interesting $\uparrow$ & Coherent $\uparrow$ & Interesting $\uparrow$ & Coherent $\uparrow$ \\
\midrule
\baby & \bf 88.3 & \bf 84.9  &  \bf 87.1 & \bf 82.0 & \bf 91.5 & \bf 82.7 \\
Rolling-ChatGPT & 13.5 & 17.1 & 14.5 & 20.1 & 11.9 & 17.7 \\
\midrule
\baby & \bf 64.1 & \bf 64.5 & \bf 66.8 & \bf 63.2 & \bf 61.0 & \bf 61.4 \\
RE$^{3}$ & 34.6 & 30.2 & 27.9 & 28.8 & 38.3 & 37.9 \\
\midrule
\baby & \bf 65.8 & \bf 60.7  & \bf 72.1 & \bf 66.8 & \bf 60.2  & \bf 58.1 \\
 DOC & 29.1 & 39.7 & 27.2 & 25.6 & 33.8 & 37.0\\
\bottomrule
\end{tabular}}
\label{tab:main}
\end{table}

%% file: tabs/tab_ablation.tex
\begin{table}[h]
\centering
\caption{Pair-wise comparison of \baby with ablated variants and the variant that uses GPT-4 as the backbone model. We sample 20 novels of different genres for comparison. Results in different comparisons are not comparable with each other. Bold indicates significance with $p < 0.05$.}
\resizebox{0.85\textwidth}{!}{
\begin{tabular}{lcccc}
\toprule
Novel genres & \multicolumn{2}{c}{Sci-Fi} & \multicolumn{2}{c}{Fantasy}  \\
$\sim$ 6000 words & Interesting $\uparrow$ & Coherent $\uparrow$ & Interesting $\uparrow$ & Coherent $\uparrow$ \\
\midrule
\baby & 58.9 & \bf 65.1  &  55.3 & \bf 64.1 \\
\ w/o Short term memory  & 44.2 & 31.0 & 47.7 & 33.5 \\
\midrule
\baby & 51.4 & \bf 71.3  &  57.5 & \bf 68.9 \\
\ w/o Long term memory & 40.0 & 27.8 & 46.2 & 38.7 \\
\midrule
\baby  & 21.3 & 28.1 & 27.1 & 24.8 \\
\ w/ GPT-4 & \bf 73.4 & \bf 64.9 & \bf 71.7 & \bf 70.5 \\
\bottomrule
\end{tabular}}
\label{tab:ablation}
\end{table}

%% file: sections/5_RelatedWorks.tex
\section{Related Works}

\subsection{Transformers Beyond Fixed-size Context} 
One major limitation of Transformers is that the context size is fixed, which hinders their ability on processing and producing long texts. Previous work attempts to solve this issue from two different ways: designing efficient attention mechanisms to train and use Transformers with larger context windows~\citep{Beltagy2020Longformer,DBLP:conf/iclr/KitaevKL20,child2019generating,DBLP:conf/nips/ZaheerGDAAOPRWY20}, and adding memory mechanisms to the computational graph in a Transformer to allow it to process information from multiple context windows~\citep{dai*2019transformerxl,wang2019rtransformer,cui-hu-2021-sliding,bulatov2022recurrent}. While these methods enable Transformers to process very long texts, they all require substantial architectural changes to the original Transformer architecture. Therefore, these approaches can not be integrated into powerful pre-trained LLMs such as ChatGPT and LLAMA, which substantially limits their usefulness. Recently, \citet{DBLP:conf/iclr/PressSL22} introduces ALiBi, which adds linear bias to attention to allow input length extrapolation. However, this method mainly supports longer inputs instead of longer outputs. In addition, it requires access to the model parameters and inference codes, which is often not possible since many state-of-the-art LLMs such as ChatGPT, GPT-4, and PaLM, are closed-sourced.

\subsection{Long Text Generation}
In addition to architectural modifications, a number of works investigate long text generation in a hierarchical manner. \citet{fan2018hierarchical} first propose to generate a story by first generating a short summary of it and then improve this method by adding an intermediate step of generating an outline which is the predicate-argument structure of the story~\citep{fan2019strategies}. \citet{tan-etal-2021-progressive} and \citet{sun-etal-2022-summarize} further improve this kind of hierarchical long text generation method.
\citet{DBLP:conf/aaai/YaoPWK0Y19} also propose to first generate a storyline and then complete the story. This line of research is further improved by RE$^3$\cite{yang-etal-2022-re3} and its variant DOC\cite{yang2022doc}, which proposed to recursively prompt LLMs for long story generation in a plan-and-write fashion. However, the plots and length of their final stories are still constrained by the pre-determined plans. In contrast, \baby overcomes the above limitations via recurrent generation, which enables effective human-LM collaboration and improves the flexibility and controllability for long text generation.

\subsection{AI-Assisted Writing Systems} 
AI writing assistants have been adopted in a variety of applications, including story completion\cite{coauthor}, essay writing~\citep{liu-etal-2019-neural-based}, and poem generation~\citep{ghazvininejad-etal-2017-hafez}. Existing systems can be broadly classified into \emph{interactive} generation and \emph{automatic} generation. Interactive systems~\citep{coenen2021wordcraft,chung2022talebrush,goldfarb2019plan} are mainly designed to provide local suggestions or revisions at the phrase or sentence level. As a result, they are less able to ease the creative burden for human writers. 
On the other hand, automatic generation ~\citep{fan2019strategies,tian-peng-2022-zero,li2013story} aims to write full texts based on given prompts or topics via the sequence-to-sequence framework. 
Although advances in LLMs have demonstrated impressive potential for these systems, the lack of transparency, controllability, and sense of collaboration could harm user experience regarding writers’ perceived ownership~\citep{coauthor,birnholtz2013write}. 
Besides, most of them are limited by providing local editing suggestions ranging from several phrases to a few sentences~\citep{han2022go,yao2019plan}, partly due to the length limitation of NLG models and partly due to the challenge of maintaining long-range coherence. 

%% file: sections/6_Limitations.tex
\section{Limitations}
One limitation of this work is that while \baby can generate arbitrarily long texts, we only evaluate it on settings where the generated texts are at most around 5000 words. This is because both qualitative and quantitive evaluations of very long texts are prohibitively hard. Another limitation is that \baby only works with backbone LLMs that are powerful enough such as ChatGPT and GPT-4. We believe this issue can be alleviated when more powerful smaller LLMs are developed. Finally, our user study for evaluating \baby as an AI writing assistant and as interactive fiction is limited by small-scale studies. We will add larger and more throughout the user study in the revised version. As for the social impact, \baby can improve the quality of AI-generated long texts and increase the productivity of human writers. However, it can also be misused to generate garbage or harmful content that leads to negative social impact. However, this is a known limitation of generative AI and we will make our best effort to promote responsible usage of generative AI.

%% file: sections/7_Conclusion.tex
\section{Conclusions}
We present \baby, a language-based simulacra of the recurrence mechanism in RNNs that uses language-based components and defines a recurrent computation graph via prompt engineering. \baby enbale LLMs to generate arbitrarily long texts either autonomously or by interacting with human writters. Its language-based components improves its interpretability and controllability and the prompt-based computation graph makes it easily customizable. User study on using \baby as AI writing assistants and text-based games demonstrates its potential as an initial step towards next-generation AI writing assistant beyond local writing suggestions and directly using generative AI as contents that are consumerable via interaction. Finally, our work also demonstates the possibility of borrowing ideas from popular model designs in cognitive science and deep learning literature for 
long form text generation using LLMs.

%% file: sections/appendix.tex
\clearpage
\begin{appendices}
\section{Prompts} \label{app:prompt}
\begin{figure}[bhtp]
    \centering
   \includegraphics[width=\textwidth,trim={0cm 0cm 0cm 0cm} ,clip]{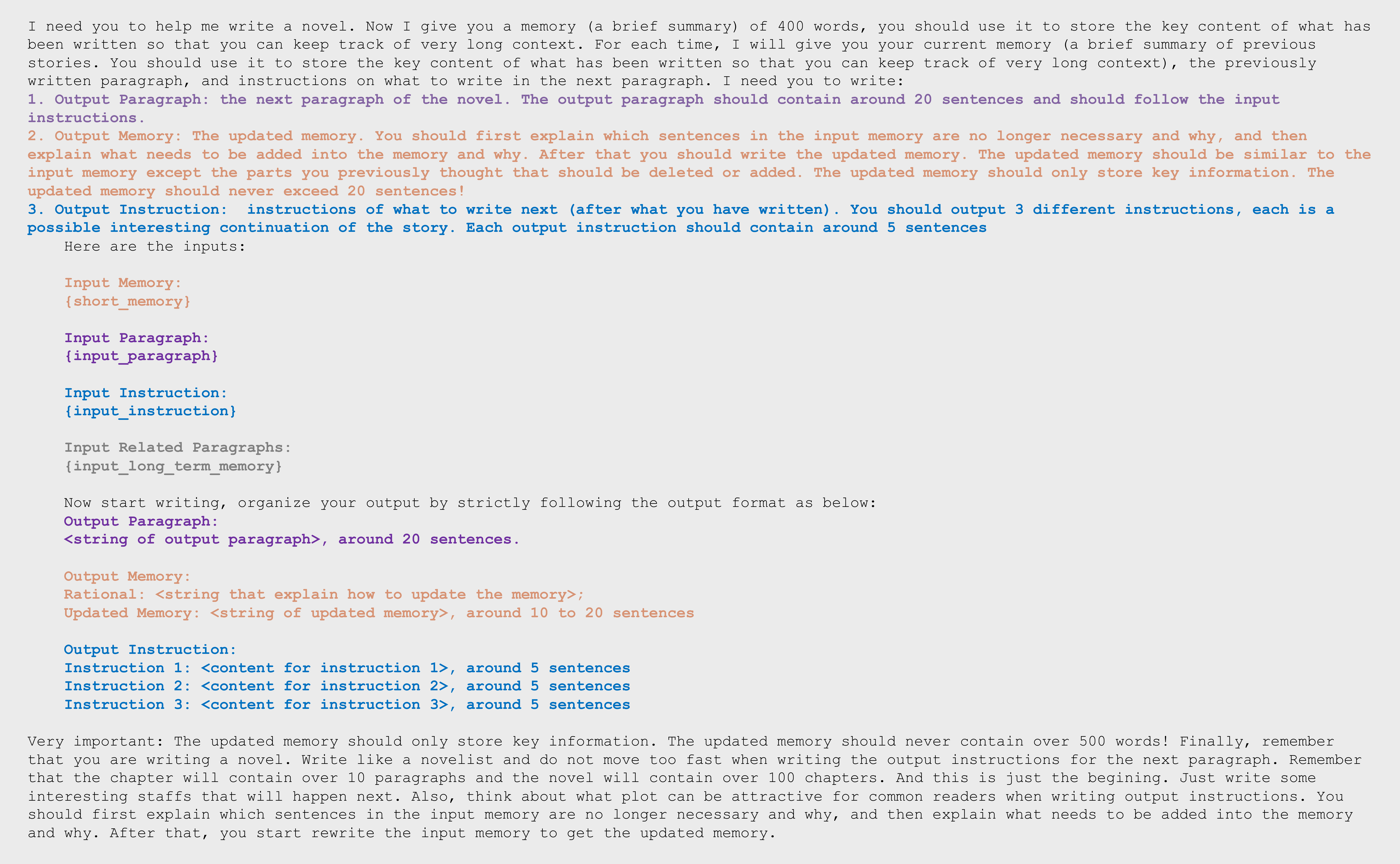}
    \caption{The prompts designed for the backbone LLM in the \baby framework that simulates input (plan, instruction), output, short-term memory, and long-term memory, respectively. \looseness=-1}
    \label{fig:prompt}
\end{figure}

\section{Demo}\label{app:demo}
\begin{figure}[bhtp]
    \centering
   \includegraphics[width=\textwidth,trim={0cm 0cm 0cm 0cm} ,clip]{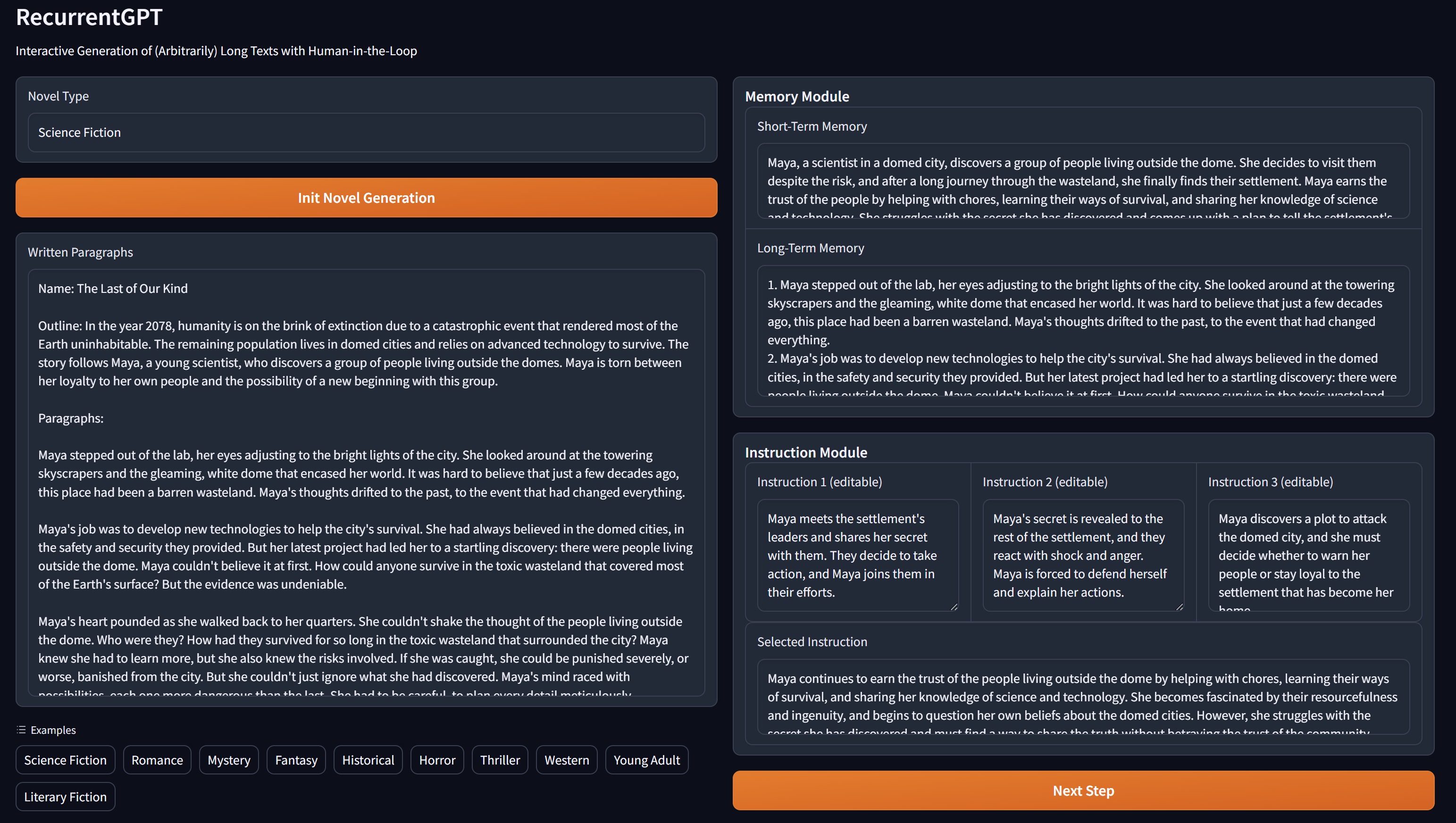}
    \caption{A web demo of \baby. \looseness=-1}
    \label{fig:demo}
\end{figure}

\end{appendices}